\documentclass[conference,10pt]{IEEEtran}

\usepackage{cite}
\usepackage{amsmath,amssymb,amsfonts}
\usepackage{algorithmic}
\usepackage{graphicx}
\usepackage{textcomp}
\usepackage{xcolor}
\def\BibTeX{{\rm B\kern-.05em{\sc i\kern-.025em b}\kern-.08em
    T\kern-.1667em\lower.7ex\hbox{E}\kern-.125emX}}

\usepackage{booktabs} 
\usepackage{soul}
\graphicspath{{figs/}}
\usepackage{braket}
\usepackage{etoolbox}
\makeatletter
\patchcmd{\@maketitle}
  {\addvspace{0.5\baselineskip}\egroup}
  {\addvspace{-0.5\baselineskip}\egroup}
  {}
  {}

\begin{document}

%\title{\LARGE{ICCAD Special Session Paper: Quantum-Classical Hybrid Machine Learning for Image Classification}
%\vspace{-4mm}}

\title{\LARGE{Quantum-Classical Hybrid Machine Learning for Image Classification \\\large{(ICCAD Special Session Paper)}}
\vspace{-4mm}}

\author{
\IEEEauthorblockN{Mahabubul~Alam$^{1}$, Satwik~Kundu$^{1}$, Rasit~Onur~Topaloglu$^{2}$, Swaroop~Ghosh$^{1}$}

\IEEEauthorblockA{
\textit{$^1$School of Electrical Engineering and Computer Science, Penn State University, University Park}\\
\textit{$^2$IBM Corporation}\\
\textit{mxa890@psu.edu, satwik@psu.edu, rasit@us.ibm.com, szg212@psu.edu}
}
}

\maketitle

\begin{abstract}

Image classification is a major application domain for conventional deep learning (DL). Quantum machine learning (QML) has the potential to revolutionize image classification. In any typical DL-based image classification, we use convolutional neural network (CNN) to extract features from the image and multi-layer perceptron network (MLP) to create the actual decision boundaries. QML models can be useful in both of these tasks. On one hand, convolution with parameterized quantum circuits (Quanvolution) can extract rich features from the images. On the other hand, quantum neural network (QNN) models can create complex decision boundaries. Therefore, Quanvolution and QNN can be used to create an end-to-end QML model for image classification. %Hybrid approaches such as, feature extraction through Quanvolution and classification with MLP, or feature extraction through CNN and classification with QNN may prove useful as well.
Alternatively, we can extract image features separately using classical dimension reduction techniques such as, Principal Components Analysis (PCA) or Convolutional Autoencoder (CAE) and use the extracted features to train a QNN. We review two proposals on quantum-classical hybrid ML models for image classification namely, Quanvolutional Neural Network and dimension reduction using a classical algorithm followed by QNN. Particularly, we make a case for trainable filters in Quanvolution and CAE-based feature extraction for image datasets (instead of dimension reduction using linear transformations such as, PCA). We discuss various design choices, potential opportunities, and drawbacks of these models. We also release a Python-based framework to create and explore these hybrid models with a variety of design choices.% for any chosen dataset. %This framework can be extended for further exploration of these models.

\end{abstract}

\section{Introduction}
%what is the current state of quantum
Quantum computing is a new computing paradigm with tremendous future potential. Even though the technology is still in a nascent stage, the community is seeking computational advantage from quantum computers (i.e., quantum supremacy) for practical applications. Recently, Google claimed quantum supremacy can be achieved even with a 53-qubit quantum processor by completing a specific computation in 200 seconds that might take 10K years \cite{arute2019quantum} (later rectified to 2.5 days \cite{pednault2019quantum}) on a state-of-the-art supercomputer. %This experiment was a significant milestone for quantum computing. %However, the computational task of this experiment had no practical value.

The near-term quantum devices have limited number of qubits. Moreover, they suffer from various types of noises (decoherence, gate errors, measurement errors, crosstalk, etc.). Due to these limitations, these machines are not yet perfectly suitable to execute quantum algorithms that rely on high orders of error correction (e.g., Shor's factorization, Grover's search). Quantum machine learning (QML) promises to achieve quantum advantage with near-term machines because it is based on a variational principle (similar to other near-term algorithms such as, Quantum Approximate Optimization Algorithm or QAOA \cite{farhi2018classification}, Variational Quantum Eigensolver or VQE \cite{kandala2017hardware} and so on) that does not necessitate error correction \cite{mcclean2016theory}.

\begin{figure*}
 \begin{center}
    \includegraphics[width=0.95\textwidth]{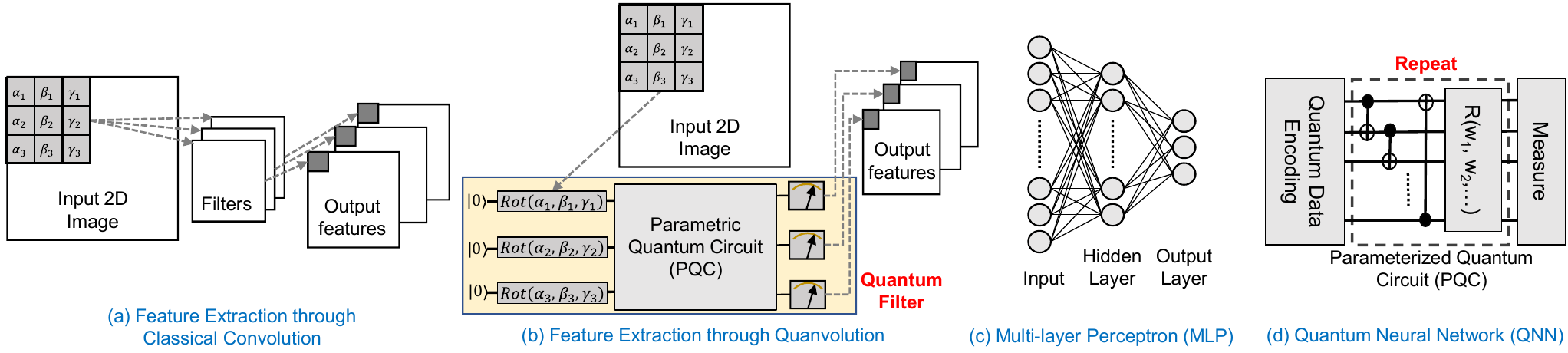}
 \end{center}
\vspace{-4mm}
\caption{
(a) shows a classical convolution operation. (b) shows a toy Quanvolution operation proposed in \cite{henderson2020quanvolutional}. In Quanvolution, a quantum circuit (also referred to as quantum filter) encodes an image segment as a quantum state, and produces output features corresponding to that segment through state transformation using a parameterized circuit and subsequent measurement operations. (c) shows the network diagram of a toy Multi-layer Perceptron (MLP) network. A conventional Quantum Neural Network (QNN) is shown in (d). It consists of a data encoding circuit, a parameterized circuit, and measurement operations.
%A conventional Quantum Neural Network (QNN) is shown in (a). It consists of a data encoding circuit, a parameterized circuit, and measurement operations. (b) and (c) show toy diagrams of Multi-layer Perceptron (MLP) network, and convolutional layer operations, respectively. They are widely used in modern deep learning applications. (d) shows a toy Quanvolution operation proposed in \cite{henderson2020quanvolutional}. In Quanvolution, a quantum circuit encodes an image segment as a quantum state, and produces output features corresponding to that segment through state transformation using a parameterized circuit and subsequent measurement operations. 
%\hl{Rasit: Since feature extraction comes first, c and d should move ahead of a and b. Since classical mentioned first in the feature extraction, QNN and MLP places should also be exchanged.}
} 
\vspace{-4mm}
\label{fig:demo}
\end{figure*}

Image classification is one of the most useful ML task having wide applications in autonomous driving \cite{chen2017multi, maturana2015voxnet}, medical diagnostics \cite{mckinney2020international, esteva2017dermatologist}, biometric security \cite{taigman2014deepface, schroff2015facenet}, to name a few. An image classification ML pipeline generally consists of two stages: (i) feature extraction, and (ii) classification based on the extracted features. Before the rise of convolutional neural networks (CNN), various statistical techniques dominated feature extraction from images (e.g., SIFT, SURF, FAST, etc.) commonly referred to as feature engineering \cite{nixon2019feature}. Later, these extracted features are used as inputs to a classifier (e.g., KNN, SVM, Decision Tree, Naive Bayes, MLP, etc.) \cite{friedman2001elements}. CNN can extract the features and learn the classification decision boundaries simultaneously, and thus, eliminates the tedious step of feature engineering. As a result, CNN has become the ML algorithm of choice for image classification in recent years. It has also achieved human-level accuracy in many image recognition tasks \cite{mckinney2020international, esteva2017dermatologist, krizhevsky2017imagenet}.  

Several QML models have been proposed for image classification to exploit quantum computers in practical use cases \cite{cong2019quantum, henderson2020quanvolutional, mari2020transfer, li2020quantum, kerenidis2019quantum, dang2018image, henderson2021methods, li2021drug}. In \cite{henderson2020quanvolutional}, the authors proposed Quanvolutional Neural Networks where parametric quantum circuits are used as filters/kernels to extract features from images. These quantum filters take image segments as inputs and produce output feature maps by transforming the data in the quantum space. %\hl{Rasit: This is the part that causes confusion; instead of saying qfilters are small QNN's, can we just say they are small quantum circuits and bound the QNN to be the decision boundary classifier definition in the context of this paper?}
The output features are used as inputs to an MLP network. In \cite{mari2020transfer}, the authors extended the classical transfer learning approach to the quantum domain. Here, the trained convolutional layers in a classical deep neural network are used to extract image features. Later, a Quantum Neural Network (QNN) is trained separately to learn the classification decision boundaries from these features. Several works have used classical dimension reduction techniques (e.g., Principal Component Analysis or PCA) to extract image features and later, used them as inputs to a QNN \cite{huang2021power, grant2018hierarchical}. In \cite{cong2019quantum}, the authors propose Quantum Convolutional Neural Network (QCNN) which is motivated by CNN. Here, convolutions are multi-qubit operations performed on neighboring pairs of qubits. These convolutions are followed by pooling layers, which are implemented by measuring a subset of the qubits, and using the results to control subsequent operations. The network ends with pairwise operations on the remaining qubits before measurement.

In this paper, we review two promising hybrid architectures for image classification, (i) Quanvolutional Neural Network (Quanvolution + MLP), and (ii) classical dimension reduction + QNN. We discuss their design choices, characteristics, enhancements, and potential drawbacks. Particularly, we advocate for trainable quantum filters in Quanvolution, and classical Convolutional Autoencoder (CAE) for image feature extraction in quantum-classical hybrid image classification models. A QNN/quantum filter has a myriad of design choices in terms of encoding methods, parametric circuits, and measurement operations. However, in this work, we only use two configurations for demonstration. The accompanying Python-based framework supports a wide variety of QNN/quantum filter design choices (6 encoding circuits, 19 parametric circuits, and 6 measurement circuits). Interested readers can utilize/extend this framework to explore the design space.

In the remaining paper, we cover basics on quantum computing and QNN in Section \ref{preli}, discuss the hybrid architectures in Section \ref{archi}, present relevant results in Section \ref{eval}, and draw the conclusions in Section \ref{conc}.

\begin{figure*}
 \begin{center}
    \includegraphics[width=0.8\textwidth]{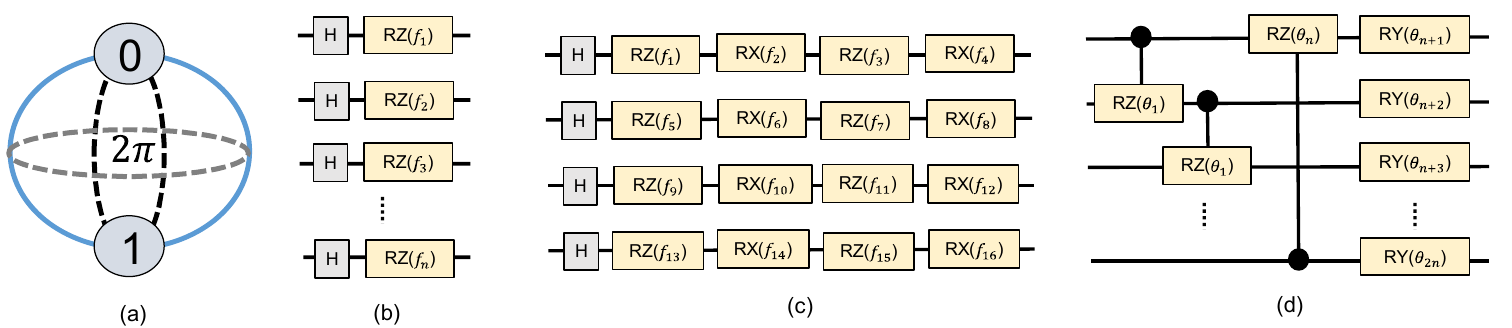}
 \end{center}
\vspace{-4mm}
\caption{ (a) Bloch sphere representation of a qubit. At any given step, a qubit can be rotated along the X, Y, or Z axis by applying a gate. The states will repeat in 2$\pi$ intervals. (b) Angle encoding 1:1 (1 continuous variable encoded in a single qubit state using RZ rotation, n qubits are required to encode n continuous variables as an n-qubit state). (c) Angle encoding 4:1 (4 continuous variables encoded in a single qubit state using alternating RZ and RX rotations, 4 qubits encode 16 continuous variables as a 4-qubit state). (d) Parametric layer used in this work. Parametric CRZ gates entangle the qubits; this is followed by single qubit RY rotations. Each n-qubit parametric layer has 2n circuit parameters.
} 
\vspace{-4mm}
\label{fig:circuit}
\end{figure*}

\section{Preliminaries} \label{preli}

\noindent {\bf{Qubits, Quantum Gates, State Vector, \& Measurements:}} Qubit is analogous to classical bits. However, unlike a classical bit, a qubit can be in a superposition state i.e., a combination of $\ket{0}$ and $\ket{1}$ at the same time. A variety of qubit technologies exists, e.g., superconducting qubits, trapped-ions, neutral atoms, silicon spin qubits, to name a few \cite{nielsen2002quantum}. Quantum gates such as, single qubit (e.g., Pauli-X ($\sigma_{x}$) gate) or multiple qubit (e.g., 2-qubit CNOT gate) gates modulate the state of qubits to perform computations. These gates can either perform a fixed or tunable computation e.g., an X gate flips a qubit state while the RY($\theta$) gate rotates the qubit along the Y-axis by $\theta$. A two-qubit gate changes the state of one qubit (\textit{target qubit}) based on the current state of the other qubit (\textit{control qubit}). %For example, the CNOT gate flips the target qubit if the control qubit is in $\ket{1}$ state. 
A quantum circuit can contain many gate operations. Qubits are measured in a desired basis to retrieve the final state of a quantum program. In physical quantum computers, measurements are generally restricted to a computational basis, e.g., Z-basis in IBM quantum computers.

%Qubit state is expressed with a \textit{ket} ($\ket{.}$) notation and the state of a set of qubits is known as \textit{state vector}. A single qubit state $\ket{\psi}$ is described as $\ket{\psi} = a \ket{0} + b \ket{1}$. Here, $\ket{0}$ and $\ket{1}$ are known as basis states represented by $[1 \hspace{2mm} 0]^T$ and $[0 \hspace{2mm} 1]^T$, and a and b are complex numbers such that $|a|^2 + |b|^2 = 1$. An n-qubit quantum state can be represented by a linear combinations of $2^n$ basis states. 

\noindent {\bf{Expectation Value of an Operator:}} Expectation value is the average of the eigenvalues, weighted by the probabilities that the state is measured to be in the corresponding eigenstate. Mathematically, expectation value of an operator ($\sigma$) is defined as $\langle\psi|\sigma|\psi\rangle$ where $|\psi\rangle$ is the qubit state vector. It varies between the minimum and maximum eigenvalues of the operator. For example, the Pauli-Z ($\sigma_z$) operator has two eigenvalues: +1 and -1. Therefore, the Pauli-Z expectation value of a qubit will vary in the range of [-1, 1] depending on the qubit state.

\noindent {\bf{Quantum Neural Network:}} QNN involves parameter optimization of a PQC to obtain a desired input-output relationship. QNN generally consists of three segments: (i) a classical to quantum data encoding (or embedding) circuit, (ii) a parameterized circuit, and (iii) measurement operations. A variety of encoding methods are available in the literature \cite{schuld2021effect}. For continuous variables, the most widely used encoding scheme is angle encoding where a continuous variable input classical feature is encoded as a rotation of a qubit along the desired axis (X/Y/Z) \cite{abbas2020power, schuld2020circuit, schuld2021effect, lloyd2020quantum}. For `n' classical features, we require `n' qubits. For example, RZ(f1) on a qubit in superposition (Hadamard - H gate is used to put the qubit in superposition) is used to encode a classical feature `f1' in Fig. \ref{fig:circuit}(b). We can also encode multiple continuous variables in a single qubit using sequential rotations. For example, `f1', `f2', `f3', and `f4' are encoded using consecutive RZ(f1), RX(f2), RZ(f2), and RX(f4) rotations on a single qubit in Fig. \ref{fig:circuit}(c). As the states produced by a qubit rotation along any axis will repeat in 2$\pi$ intervals (Fig. \ref{fig:circuit}(a)), features are generally scaled within 0 to 2$\pi$ (or -$\pi$ to $\pi$) in a data pre-processing step. %Discrete/categorical variables can be encoded using rotations in discrete steps \cite{yano2020efficient}.

The parametric circuit has two components: entangling operations and parameterized single-qubit rotations. The entanglement operations are a set of multi-qubit operations between all the qubits to generate correlated states \cite{lloyd2020quantum}. The following parametric single-qubit operations are used to search through the solution space. This combination of entangling and single-qubit rotation operations is referred to as a parametric layer in QNN. 
%Note that significant research has been done in the past few years on finding the optimal PQC architecture for QNN. Descriptors such as, expressive power, entanglement capability, effective dimension, etc. have been proposed to measure the potency of various PQC choices \cite{sim2019expressibility, abbas2020power, du2020expressive}. The proponents of these descriptors claim significant correlation between the descriptor values and the trainability of the quantum circuits. In practical applications, such descriptors may be useful to choose optimal PQC architecture for intended QML application. 
A widely used parametric layer architecture is shown in Fig. \ref{fig:circuit}(d) \cite{sim2019expressibility, abbas2020power}. Here, CRZ($\theta$) gates between neighboring qubits create the entanglement, which is followed by rotations along Y-axis using RY($\theta$). %\hl{Rasit: Something seems wrong here because the figure does not contain CNOTs}.
Normally, these layers are repeated multiple times to extend the search space \cite{schuld2020circuit, abbas2020power}.

\noindent {\bf{QNN Cost Functions:}} Qubits in a QNN circuit are measured in the computational basis to retrieve the output state. A cost function is derived from the measurements to train the network \cite{abbas2020power, schuld2019quantum, schuld2020circuit}. For example, in a binary classification problem, the authors measured all the qubits in the QNN model in Pauli-Z basis and associated class 0 with the probability of obtaining even parity, and class 1 with odd parity \cite{abbas2020power}. Then, the model is trained using binary cross-entropy loss. In \cite{alam2019addressing}, the authors used the Pauli-Z expectation value of a single qubit (-1 associated with class 1 and +1 associated with class 0) for a binary classifier and trained it using mean squared error (MSE) loss.% \hl{Rasit: MSE needs definition}. %Note that, multiple one-vs-all classifiers are often used for multi-class classification problems (i.e., classification of more than two classes) \citealp{schuld2020circuit}. 
In \cite{huang2021power}, the authors fed the outputs of the QNN to a classical neural network and trained it using the binary cross-entropy loss function.% (Projected Quantum Kernel or PQK).

\noindent {\bf{Training QNN:}}
QNN's can be trained using any gradient-based optimization algorithm such as, Adam \cite{kingma2014adam} or Adagrad \cite{duchi2011adaptive}. To apply these methods, we need to compute the gradients \cite{banchi2021measuring, schuld2019evaluating} of the QNN outputs with respect to the circuit parameters. 
%Multiple methods exist to compute these gradients \cite{luo2020yao, banchi2021measuring, schuld2019evaluating}. 
The parameter-shift rule is a known method to compute the gradients \cite{banchi2021measuring, schuld2019evaluating}. Conceptually, parameter-shift rule is very similar to the age-old finite difference method which uses two evaluations of a target function at close proximity to compute the gradients with respect to a parameter. Unlike finite difference, the two data points can be far from each other in parameter-shift rule. As a result, it shows greater resilience to shot noise and measurement errors compared to finite difference \cite{schuld2019evaluating}. 
%The PennyLane framework provides support for many quantum gradient computation methods \cite{bergholm2018pennylane}. 
Alternatively, one can also use gradient free optimizer such as Nelder-Mead to train a QNN \cite{lavrijsen2020classical}. However, a gradient-free optimizer may perform poorly when the network has lots of parameters.

\section{Hybrid Architectures for Image Classification} \label{archi}

\subsection{Quanvolution + MLP}

Quanvolution is simply an extension of classical convolution
%. The convolutional neural network (CNN) is based on the idea of a convolution layer where, instead of processing the full input image data with a global function, we apply local convolution. 
where small local regions of an image are sequentially processed with the same kernel/filter. A kernel is a small 2D matrix. The dot product between the kernel and the image segment is used to generate an output feature. %The number of computations required is simply $O(n^2)$ 
%\hl{Rasit: n needs defined. Also, shouldn't it depend on kernel size also?}. 
For 3D RGB images, separate kernels are applied across the channels (2D planes), and they are collectively referred to as filters. For 2D images, filters and kernels are synonymous. The results obtained for each region are usually associated to different channels of a single output pixel. The union of all the output pixels produces a new image-like object, which can be further processed by additional layers. A toy convolutional layer operation is shown in Fig. \ref{fig:demo}(a). In Quanvolution, quantum circuits mimic the behavior of classical CNN filters.

\begin{figure}
 \begin{center}
    \includegraphics[width=0.3\textwidth]{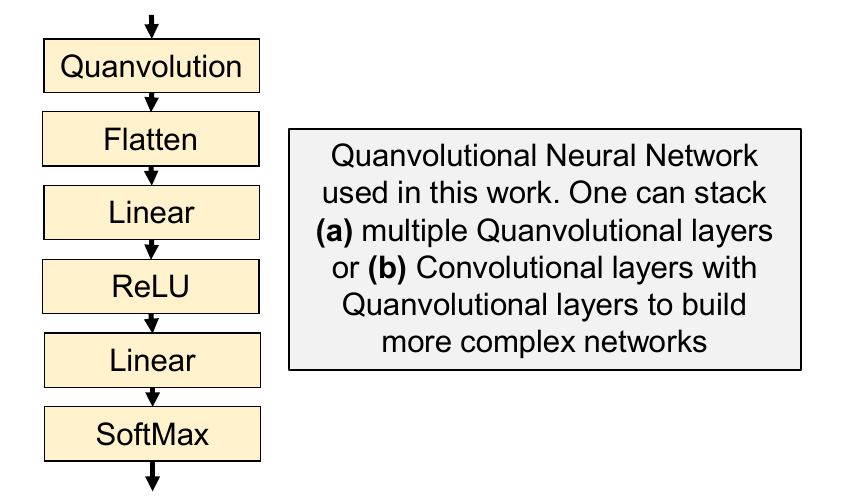}
 \end{center}
\vspace{-4mm}
\caption{A quantum-classical hybrid neural network based on the Quanvolutional Neural Network architecture \cite{henderson2020quanvolutional} with a single Quanvolutional layer.
}
\vspace{-6mm}
\label{fig:quanv}
\end{figure}

\noindent {\bf{Quantum Filters:}}
The quantum filters encode image segments as input state of a quantum circuit. The state is transformed using a parameterized quantum circuit and subsequent measurement operations produce output features corresponding to that segment. Fig. \ref{fig:demo}(b) shows a toy Quanvolution layer. Here, a 3-qubit quantum circuit is used as a filter. It encodes 3x3 image segments as a 3-qubit quantum state using 3 Rot($\alpha,\beta,\gamma$) rotations (Rot is an arbitrary quantum gate that takes three rotation parameters). Similar to CNN, these quantum filters are moved across the 2D plane in finite steps (strides) to generate the complete output feature map of the image.

The qubit size of a filter depends on the chosen encoding method and the kernel size. For example, if we use the 1 variable/qubit encoding method (Fig. \ref{fig:circuit}(b)), the resulting quantum filter will be a 16-qubit circuit for kernel size of 4x4. It will reduce to a 4-qubit circuit if we choose the 4 variables/qubit encoding method (Fig. \ref{fig:circuit}(c)). Concepts like data re-uploading \cite{perez2020data} can be used to encode an arbitrary number of variables in an arbitrary number of qubits. Choice of encoding methods will most likely be dictated by the availability of quantum resources \cite{henderson2021methods}. In this work, we use the 4 variables/qubit encoding method for 4x4 kernels. We use the circuit architecture shown in Fig. \ref{fig:circuit}(d) (Circuit 13 of \cite{sim2019expressibility}) as our preferred PQC in the quantum filters. We also use the Pauli-Z expectation values of the qubits as output features (an n-qubit filter generates n-features/image segment). Increasing the number of filters increases the number of extracted features for the downstream classifier. Similar to a CNN with high number of stages, a large number of quantum filters improve performance (lower cost/higher accuracy/faster training) \cite{henderson2020quanvolutional}. 

%\hl{isn't following two para repeated?}

%The encoding circuit is followed by a PQC. However, numerous choices exist for PQC construction. The work in \cite{sim2019expressibility} summarizes 19 widely used PQC architectures in contemporary QNN literature. In this work, we use the PQC architecture shown in Fig. \ref{fig:circuit}(d)
%(Circuit 13 of \cite{sim2019expressibility}). This PQC 
%which consists of qubit pair-wise parametric CRZ rotations (controlled rotation along the Z-axis) followed by single-qubit RY rotations. We also restrict the number of parametric layers in the filter to 3.

%The output state measurement operations in the filters generate the output features. Again, there is a wide variety of choices for the task. For example, one can use the probabilities of different basis state measurements as the output features (an n-qubit filter will generate $2^n$ features for each image segments). Alternatively, one can use only the probability of a single basis state (e.g., all-zero state) as the output feature (an n-qubit filter will generate a single feature for each image segments). In this work, we use the Pauli-Z expectation values of the qubits as output features (an n-qubit filter generates n output features for each image segments).

\noindent {\bf{Filter Trainability:}} 
In the original work in \cite{henderson2020quanvolutional}, the quantum filters did not have any trainable parameters. However, in CNN, filters have trainable weights, and they are learned during the training. Similarly, quantum filters can too have trainable parameters. For example, we can either initialize the PQC parameters ($\theta_1-\theta_{2n}$ in Fig. \ref{fig:circuit}(d)) randomly and keep it constant throughout the training, or we can update them during the training alongside other parameters in the network. 

Trainable filters will result in many-fold increase in quantum circuit execution during the training. For example, if a quantum filter has p trainable parameters, it will add 2xp more quantum circuit executions for each image segment to compute the required gradients using parameter-shift rule \cite{schuld2019evaluating}. Note that, feature generation using non-trainable quantum filters is equivalent to random transformation of the image segments. A classical random function can replace such filters. In fact, the work in \cite{henderson2020quanvolutional} showed that the performance of classical random transformations of the image segments matches the performance of random transformations with non-trainable quantum filters. 
%Therefore, there is little or no motivation to use non-trainable quantum filters in practical applications- then why should we use trainable filter?}. 
However, the trainable quantum circuits are hard to simulate classically \cite{havlivcek2019supervised}. If they exhibit significant performance benefits over their non-trainable counterparts, it will be worthwhile for the research community to explore them for possible quantum advantage \cite{atchade2021quantum}.% \hl{Rasit: Missing reference, Quantum Enhanced Filter: QFilter, Atchade-Adelomou and Alonso-Linaje}

\noindent {\bf{Network Design:}}
Similar to CNN, a Quanvolutional layer can have many filters and multiple Quanvolutional layers can be stacked upon each other to develop a deep Quanvolutional Neural Network \cite{henderson2020quanvolutional, henderson2021methods}. The outputs from the final Quanvolutional layer can be fed to a an MLP (or a QNN). One can also apply classical non-linear activation functions (for additional non-linearity) and maxpooling (downsampling) at the output of a Quanvolutional layer. One can create separate filters by initializing the same PQC with different random seeds. Alternatively, we can use different PQC architectures, encoding methods, and measurement operations altogether to create different filters. One can also stack classical convolutional layers with Quanvolutional layers. Fig. \ref{fig:quanv} shows the network diagram used in this work with a single Quanvolutional layer followed by two fully connected classical layers.

\noindent {\bf{Number of Circuit Executions:}}
The number of quantum circuit executions per sample during training/inference depends on the kernel size, image size, and the stride (the amount of movement of the kernel in terms of pixels). For a 28x28 image and 4x4 kernel size, we need to execute a total of 7x7 quantum circuits when stride = 4 (a single non-trainable quantum filter). If this filter has 10 trainable parameters, the total number of circuit execution becomes 7x7 + 2x10x7x7 where the later 2x10x7x7 circuit executions are necessary to compute the gradients using parameter-shift rule (2 extra circuit executions for each parameters \cite{schuld2019evaluating}). If we take 50 samples per batch, we will need 50x(7x7 + 2x10x7x7) filter execution for each batches during training. %\hl{Rasit: Two different multiplication signs used. Please explain the formula, why do we add and why do we multiply by 2 for example} 
This is in fact a prohibitively large number for a single batch and a filter. However, all these circuits are independent of each other. Hence, one can argue that all these computations can be done simultaneously if one has access to multiple quantum computing resources.

%\noindent {\bf{Impact of Number of Filters:}}
%Increasing the number of filters essentially increases the number of extracted features for the downstream classifier network. Similar to CNN, a large number of filters improves learning (lower cost, higher accuracy, and faster training) \cite{henderson2020quanvolutional, henderson2021methods} at the cost of higher number of quantum circuit execution during both training and inference.

\subsection{Classical Dimension Reduction + QNN}

Another popular hybrid QML model targeted for smaller quantum devices uses classical algorithm (e.g., Principal Component Analysis or PCA, Linear Discriminant Analysis or LDA, etc.) to reduce data dimension to a level that is tractable for a small QNN model \cite{grant2018hierarchical, batra2021quantum, huang2021power}. Although PCA/LDA work quite well to extract most salient features from small tabular data, they are not suitable to extract features from large images. Autoencoders (AE), particularly Convolutional Autoencoders (CAE) are much more powerful tools for image feature extraction/dimension reduction \cite{masci2011stacked, chen2017deep}. PCA is a linear transformation of the data whereas AE/CAE can model more complex non-linear relationships in the data using non-linear activation functions and regularization \cite{chen2017deep}.

\begin{figure}
 \begin{center}
    \includegraphics[width=0.5\textwidth]{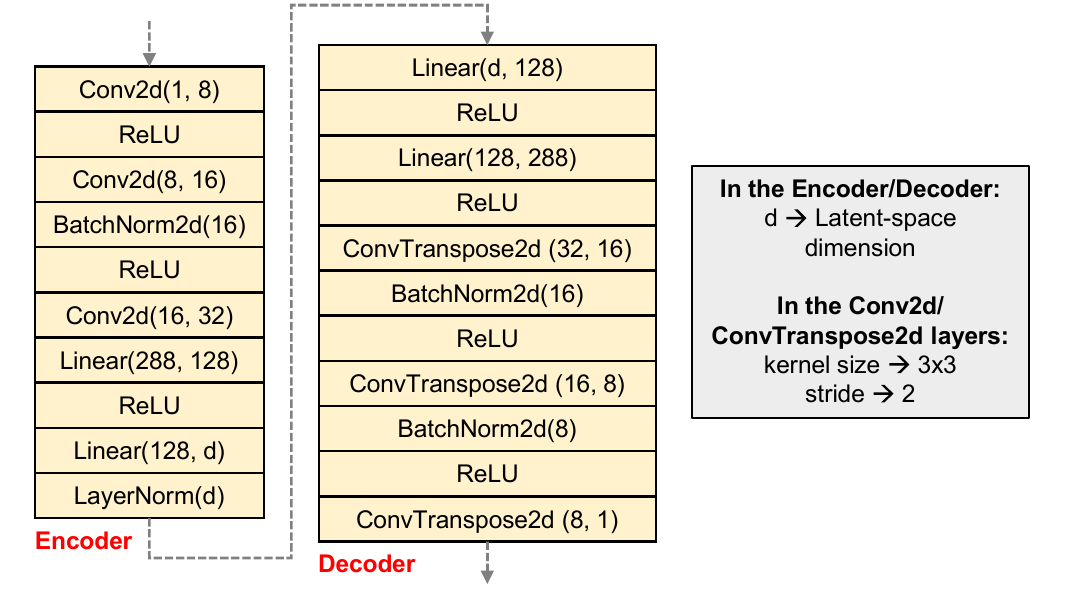}
 \end{center}
\vspace{-4mm}
\caption{Convolutional Autoencoder (CAE) architecture used in this work to extract image features (for both the MNIST and Fashion-MNIST datasets).} 
\vspace{-4mm}
\label{fig:cae}
\end{figure}

\noindent {\bf{AE/CAE:}}
AE's are a specific type of feedforward neural networks. %where the input is the same as the output. %\hl{Rasit: do you mean same size? -- Same size, same values: Mahabubul}. 
They compress the input into a lower-dimensional code using an encoder network and then reconstruct the output from this representation through a decoder network. The code is a compact representation of the input, also called the latent-space representation. The distance between the input and reconstructed output (e.g., MSE loss) is used as the feedback signal to train the network. Both the encoder and decoder in a simple AE consist of several fully connected layers. CAE provides a better architecture than AE to extract the textural features of images. In CAE, the encoder block starts with one or more successive convolutional layers. The decoder block ends with convolutional transpose/deconvolutional layers. In the middle there is a fully connected AE whose innermost layer is composed of a small number of neurons. Once trained, the encoder block can be used as a standalone entity to extract lower dimensional representation of the input data. 

Fig. \ref{fig:cae} shows the CAE network architecture used in this work (for both MNIST and Fashion-MNIST datasets). The final ConvTranspose2d layer uses Sigmoid activation where `d' is the dimension of the latent-space.

\begin{figure}
 \begin{center}
    \includegraphics[width=0.4\textwidth]{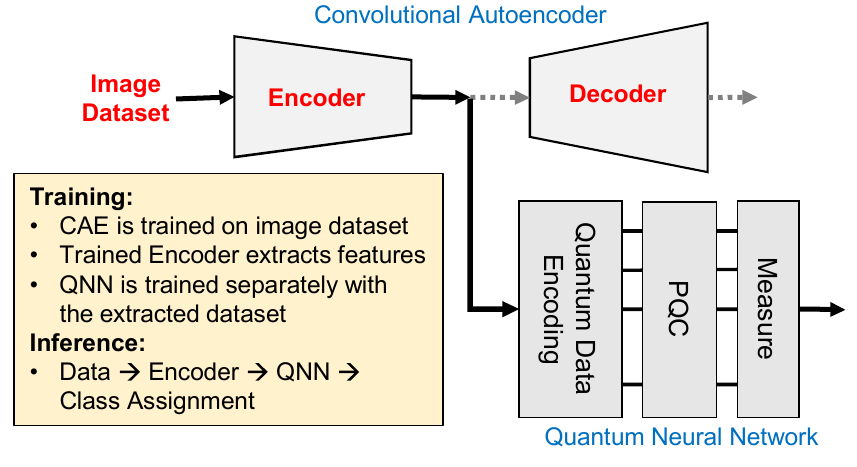}
 \end{center}
\vspace{-4mm}
\caption{The CAE + QNN network architecture. The trained CAE Encoder block creates a lower-dimensional representation of the image for the QNN.} 
\label{fig:cae_qnn}
\vspace{-4mm}
\end{figure}

\noindent {\bf{Network Design:}}
The hybrid network (Fig. \ref{fig:cae_qnn}) consists of two separate networks - a CAE and a QNN. The CAE is trained with the original image dataset to learn a lower dimensional representation of the data. The trained encoder network is used to extract image features. A conventional QNN is trained with these extracted features and image labels to perform final classification. 
%A cartoon of the network diagram is shown in Fig. \ref{fig:cae_qnn}. 
When both of these networks are trained, the encoder block and the QNN block is used together to classify data samples. We refer to this architecture as CAE+QNN.

\noindent {\bf{QNN Design-Space:}}
As mentioned earlier, numerous choices exist for the encoding circuits, PQC, and measurement circuits to build a QNN model. The accompanying Python framework to this work supports a wide variety of these choices which will impact the learnability of the QNN \cite{sim2019expressibility}. However, in this work we only use the single feature/qubit encoding method (Fig. \ref{fig:circuit}(b)), the PQC layer of Fig. \ref{fig:circuit}(d), and Z-basis measurements of the qubits in the QNN. We also restrict the number of parametric layers to 3. Following \cite{huang2021power}, we feed the QNN outputs to a fully-connected layer. The number of output neurons is equal to the number of classes in the dataset.

\noindent {\bf{CAE+QNN Vs. Transfer Learning:}}
Although, the CAE+QNN network has some similarities with transfer learning \cite{mari2020transfer}, there are some noteworthy differences as well. Both of these approaches extract image features using a classical network. In transfer learning, the convolutional layers of a classical CNN network, trained with a different dataset (e.g., AlexNet trained to classify the ImageNet dataset), is used to extract features for a target dataset. In contrast, the CAE in CAE+QNN network is trained separately to extract features from the target dataset. Therefore, the CAE extracted features may capture more variance in the target dataset compared to transfer learning, and thus, it may provide better performance (lower training cost/higher accuracy). The features extracted through transfer learning can be more generic \cite{mari2020transfer}, and it eliminates the need to train the classical network separately (as it is already trained).

\begin{table*}
\centering
\caption{Quanvolutional neural network performance after 10 epochs of training (Optimizer: Adagrad, learning rate: 0.5)% on various datasets with a non-trainable and a trainable quantum filter.% On average, quanvolution with trainable filter provides lower training/validation loss and higher training/validation accuracy over quanvolution with non-trainable filter.
} 
\label{tab:quanv_table}
\vspace{-2mm}
\includegraphics[scale=0.75]{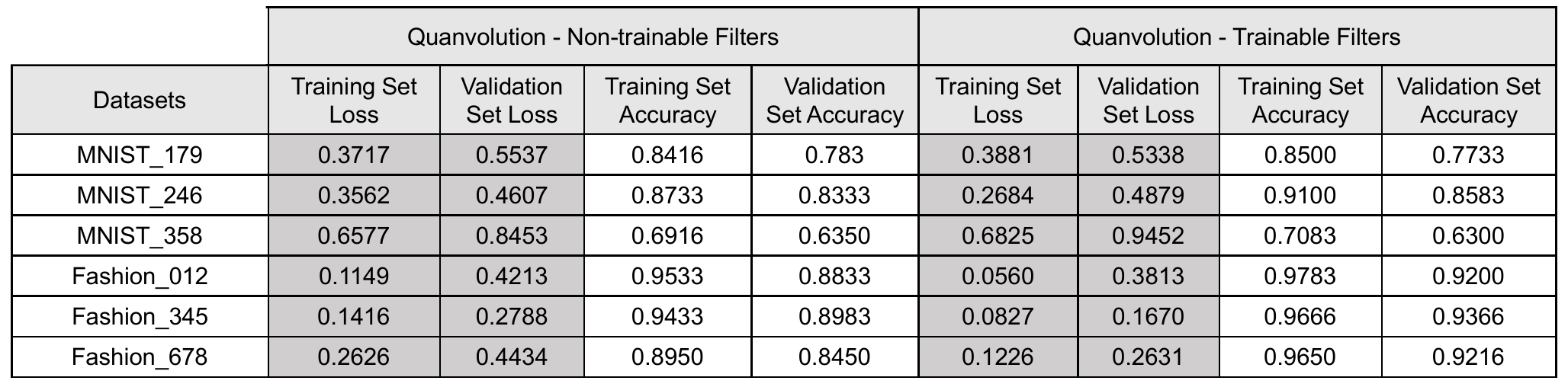}
\vspace{-4mm}
\end{table*}

\begin{table*}
\centering
\caption{CAE + QNN network performance after 20 epochs of training of the QNN (Optimizer: SGD, learning rate: 0.5)% on various datasets with latent-space dimension of 5 and 10.% When latent-space dimension = 5 (10), the CAE Encoder extracts 5 (10) data features which is fed to a 5 (10) qubit QNN. Increasing latent-space dimension generally improves performance.
} 
\label{tab:cae_qnn_table}
\vspace{-2mm}
\includegraphics[scale=0.75]{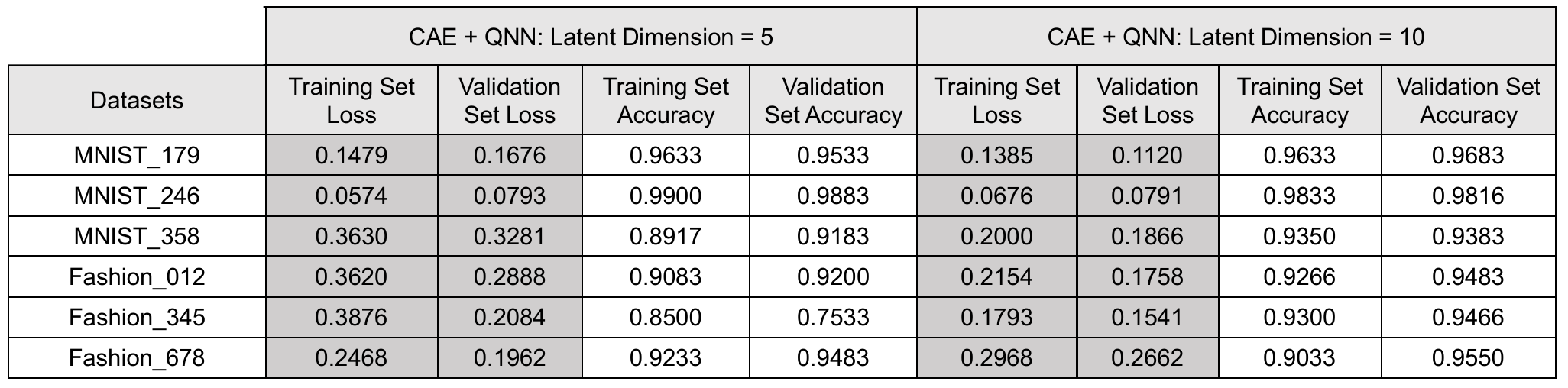}
\vspace{-4mm}
\end{table*}

\section{Evaluation} \label{eval}

In this Section, we compare the performance differences between (a) Quanvolutional Neural Networks with trainable filters, and non-trainable filters, and (b) CAE + QNN and PCA-based approach for a variety of datasets.

\noindent {\bf{Datasets:}}
We pick the MNIST and Fashion-MNIST datasets for this work which are widely used in contemporary works in QML research (each pixel value scaled within 0-1) \cite{deng2012mnist, xiao2017fashion}. Both of these datasets have 60,000 training samples, and 10,000 test samples of 2D images (28x28 pixels) that belong to 10 different classes. For empirical evaluation of the Quanvolution approach, we have picked 1,200 samples from three different classes to create 6 smaller classification datasets - MNIST\_179, MNIST\_246, MNIST\_358, Fashion\_012, Fashion\_345, and Fashion\_678. MNIST\_179 has $\approx$400 samples of digits 1, 7, and 9 each. Similarly, Fashion\_012 has $\approx$400 samples of classes 0 (t-shirt/top), 1 (trouser/pant), and 2 (pullover shirt) each. We have reduced the dimension of the samples from 28x28 to 14x14 using maxpooling to lower the simulation time. To compare CAE and PCA, we have trained the corresponding CAE (with latent dimension of 5 and 10) and PCA models with the entire MNIST and Fashion-MNIST datasets. Later, we have created 6 smaller classification datasets as before using the trained models with various latent dimensions (5/10) and principal component (10).

\noindent {\bf{Metrics:}}
We have divided the datasets into two equal sets for training and validation (600 samples/set). We use the average loss and accuracy over the entire training and validation datasets to measure the performance of the QML models \cite{anthony2009neural}. %The training loss and accuracy indicate the trainability of the models while the validation loss and accuracy indicate their generalization capability.

\noindent {\bf{Training Setup:}}
We use the gradient-based Adagrad, SGD, and Adam optimizers to train these models \cite{paszke2019pytorch, abadi2016tensorflow}. We use the same set of hyper-parameters across all the runs (learning rate = 0.5 for all the quantum/hybrid models).

%\noindent {\bf{Quanvolutional Filters and QNN:}}
%We use 4-qubit circuits as Quanvolutional filters. We use the 4 variables/qubit encoding method shown in Fig. \ref{fig:circuit}(c) to encode 4x4 pixels into a 4-qubit state. The input pixels are scaled to 0-$2\pi$ (originally 0-1). The PQC architecture in Fig. \ref{fig:circuit}(d) is used with 3 parametric layers. In Quanvolution with trainable filters, these PQC parameters are trained alongside other network parameters using gradient descent. In the non-trainable version of the filter, we set the PQC parameters randomly (-$\pi$ to $\pi$) at the beginning, and keep them constant throughout the training. Qubit Pauli-Z expectation values are used as the output features.  

\noindent {\bf{Trainable Vs. Non-trainable Filters in Quanvolution:}}
We trained quanvolutional neural networks with a single quanvolutional layer (Fig. \ref{fig:quanv}) for six 3-class classification problems with a trainable and a non-trainable quantum filter (stride = 4). We used 4-qubit circuits as Quanvolutional filters. We used the 4 variables/qubit encoding method shown in Fig. \ref{fig:circuit}(c) to encode 4x4 pixels into a 4-qubit state. The input pixels were scaled to 0-$2\pi$ (originally 0-1). The PQC architecture in Fig. \ref{fig:circuit}(d) was used with 3 parametric layers (3x2x4 parameters). In Quanvolution with trainable filters, these PQC parameters were trained alongside other network parameters using gradient descent. In the non-trainable filter, we had set the PQC parameters randomly (-$\pi$ to $\pi$) at the beginning, and kept them constant throughout the training. Pauli-Z expectation values of qubits were used as the output features. The results are tabulated in Table \ref{tab:quanv_table} (performance after 10 training epochs). Each training epoch took $\approx$195 seconds with the trainable filter compared to $\approx$57 with the non-trainable filter on a single Core i7-10750H machine with 16 GB RAM. %\hl{Rasit: Are you able to report some runtimes? Maybe some more details are needed, for example, how many parameters in total used for PQC, any other counts so that one could try to replicate your experiment?} 

On average, Quanvolution with trainable filter provided 15.98\% lower training loss, 7.49\% lower validation loss, 3.46\% higher training accuracy, and 3.32\% higher validation accuracy after 10 training epochs. In some cases, the Quanvolution with non-trainable filter performed at a similar level as its trainable counterpart. For example, MNIST\_179 and MNIST\_358 provided similar performance in both these approaches. However, in all other cases, there was a noticeable performance gap between these two approaches. We repeated the experiments 5 times with the MNIST\_179 and MNIST\_358 dataset with different random initialization. However, the performance remained at similar levels in both these approaches. In fact, both these models performed poorly on these two datasets compared to the others (average training loss of 0.535 against 0.266 overall with the trainable filter). The overall results indicate potential benefits of trainable filters which can be worthwhile to explore in the future. %\hl{Rasit: MNIST-179 and MNIST-358 are worst for trainable filter, but this is not expected. Also, why is Fashion more suitable for trainable filters?}

\noindent {\bf{CAE + QNN:}}
We trained the CAE in Fig. \ref{fig:cae} with 60000 training samples of the MNIST and Fashion-MNIST datasets with latent-dimension of 5 and 10 (optimizer: Adam, learning rate: 0.001, weight-decay: $e^{-5}$, epochs: 30, batch-size: 50). The extracted datasets (5/10 features) were used to train a QNN. In the QNN, we used 1 variable/qubit encoding method as shown in Fig. \ref{fig:circuit}(b). We used 5 and 10 qubits for the 5-feature and 10-feature datasets, respectively. The QNN shared same PQC architectures and output measurements as the quantum filters (Fig. \ref{fig:circuit}(d)). We restricted the parametric layers to 3 in the 10-qubit model (3x2x10 parameters). To match the number of trainable circuit parameters, we restricted the parametric layers to 6 in the 5-qubit models (6x2x5 parameters).

The results are tabulated in Table \ref{tab:cae_qnn_table} (performance after 20 epochs of training). All these models were trainable as evident from their loss and accuracy values. In the CAE+QNN model, the chosen number of latent-dimension (d) dictates the QNN architecture. It also affects the overall network performance. A higher value of d means more input features for the QNN model that generally translates to better training performance of the QNN. On average, the CAE + QNN model with d = 10 provided 29.85\% lower training loss, 23.23\% lower validation loss, 2.08\% higher training accuracy, and 4.68\% higher validation accuracy after 20 training epochs compared to d = 5. Therefore, a higher d (at the cost of larger QNN) may provide better performance in practical applications. %Note that, with a higher d, we also require a .

\noindent {\bf{CAE + QNN Vs. PCA + QNN:}} 
%Later, we compare the performance of CAE+QNN with PCA+QNN. 
As PCA uses linear transformation, the extracted image features are expected to be poor which may translate to poor training performance of the QNN. To perform this comparison, we trained a PCA model with the 60000 MNIST training samples and extracted 4000 samples ($\approx$400/class) as before with 10 principal components as the feature variables. We also extracted another 4000 samples from trained CAE with d = 10. Later, we trained 10-qubit QNN models (parametric layers set to 3) with these datasets for 20 epochs using the same set of training hyper-parameters (optimizer: Adagrad, learning rate: 0.5). The results are shown in Table \ref{tab:pca_cae_table}. As expected, the CAE+QNN approach outperformed the PCA+QNN approach by significant margin. The CAE + QNN model provided 48.47\% lower training loss, 49.2\% lower validation loss, 14.1\% higher training accuracy, and 25.3\% higher validation accuracy.

\begin{table}
\centering
\caption{CAE + QNN and PCA + QNN network performance after 20 epochs of training % (Optimizer: Adagrad, learning rate: 0.5) 
on MNIST dataset (4000 samples, 10 classes)
} 
\label{tab:pca_cae_table}
\vspace{-2mm}
\includegraphics[scale=0.8]{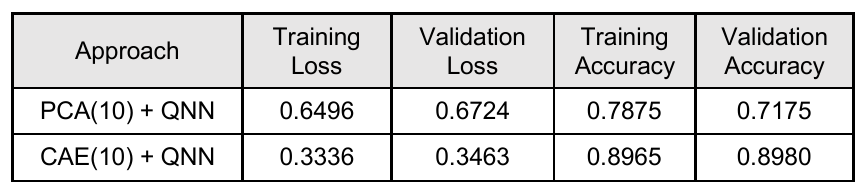}
\vspace{-6mm}
\end{table}

\noindent {\bf{Python Framework Supports:}}
Numerous choices exist for the QNN/quantum filter design using various encoding methods, parametric circuit architectures, and measurements. In this work, we only explored a limited set (Fig. \ref{fig:circuit}). However, we also release a Python-based framework (created using PennyLane, TensorFlow, and PyTorch packages \cite{bergholm2018pennylane, abadi2016tensorflow, paszke2019pytorch}) that supports 19 parametric circuit architectures from \cite{sim2019expressibility}, 6 encoding techniques, and 6 measurement circuits \cite{repo}. %\footnote{https://github.com/mahabubul-alam/iccad\_2021\_invited\_QML}. 
We also make the datasets available through the repository. Interested readers can utilize this repository for further exploration of these models on any chosen dataset.

\section{Conclusion} \label{conc}

In this article, 
%we review two promising quantum-classical hybrid machine learning architectures for image classification - Quanvolutional Neural Network and Classical Dimension Reduction + Quantum Neural Network. With empirical evidence, 
using empirical evidence we argue that trainable quantum filters in Quanvolution may provide performance benefits over the non-trainable filters, and thus, it can be worthwhile to explore for potential quantum advantage on image classification tasks. We also show that in the later architecture, dimension reduction with convolutional autoencoder (CAE) can be more useful compared to the linear transformation-based approaches such as, PCA for image datasets.   

\textbf{Acknowledgements:} The work is supported in parts by NSF (CNS-1722557, CCF-1718474, OIA-2040667, DGE-1723687 and DGE-1821766) and seed grants from Penn State ICDS and Huck Institute of the Life Sciences.

\bibliographystyle{IEEEtran}
\bibliography{ref}

\end{document}